\documentclass[pdflatex, referee, sn-basic, Numbered]{sn-jnl}

\usepackage{amsmath}
\usepackage{amssymb}
\usepackage{tabularx}
\usepackage{booktabs}
\usepackage{multirow}
\usepackage{array}

\usepackage{subcaption}

\usepackage{graphicx}
\graphicspath{ {figures/} }

\usepackage{xcolor}
\usepackage{xurl}
\usepackage{manyfoot}
\usepackage{bookmark}

\begin{document}

\title[Improving Performance of Automatic Keyword Extraction (AKE) Methods Using PoS-Tagging and Enhanced Semantic-Awareness]{Improving Performance of Automatic Keyword Extraction (AKE) Methods Using PoS-Tagging and Enhanced Semantic-Awareness}

\author*[1]{\fnm{Enes} \sur{Altuncu}}
\email{ea483@kent.ac.uk}
\author[1]{\fnm{Jason R.C.} \sur{Nurse}}
\email{J.R.C.Nurse@kent.ac.uk}
\author[2]{\fnm{Yang} \sur{Xu}}
\email{xuyang2018@sjtu.edu.cn}
\author[2]{\fnm{Jie} \sur{Guo}}
\email{guojie@sjtu.edu.cn}
\author*[1]{\fnm{Shujun} \sur{Li}}
\email{S.J.Li@kent.ac.uk}

\affil[1]{\orgdiv{Institute of Cyber Security for Society (iCSS) \& School of Computing}, \orgname{University of Kent}, \city{Canterbury}, \postcode{CT2 7NP}, \country{UK}}

\affil[2]{\orgname{Shanghai Jiao Tong University}, \city{Shanghai}, \country{China}}

\abstract{
Automatic keyword extraction (AKE) has gained more importance with the increasing amount of digital textual data that modern computing systems process. It has various applications in information retrieval (IR) and natural language processing (NLP), including text summarisation, topic analysis and document indexing. This paper proposes a \emph{simple} but \emph{effective} post-processing-based universal approach to improving the performance of \emph{any} AKE methods, via an enhanced level of semantic-awareness supported by PoS-tagging. To demonstrate the performance of the proposed approach, we considered word types retrieved from a PoS-tagging step and two representative sources of semantic information -- specialised terms defined in one or more context-dependent thesauri, and named entities in Wikipedia. The above three steps can be simply added to the end of \emph{any} AKE methods as part of a post-processor, which simply re-evaluates all candidate keywords following some \emph{context-specific} and \emph{semantic-aware} criteria. For five state-of-the-art (SOTA) AKE methods, our experimental results with 17 selected datasets showed that the proposed approach improved their performances both \emph{consistently} (up to 100\% in terms of improved cases) and \emph{significantly} (between 10.2\% and 53.8\%, with an average of 25.8\%, in terms of F1-score and across all five methods), especially when all the three enhancement steps are used. Our results have profound implications considering the fact that our proposed approach can be easily applied to any AKE method with the standard output (candidate keywords and scores) and the ease to further extend it.
}


\maketitle

\section{Introduction}

Keyword extraction (KE), also known as keyphrase or key term extraction, is an information extraction task that aims to identify a number of words/phrases that best summarise the nature or the context of a piece of text. It has several applications in information retrieval (IR) and natural language processing (NLP), including text summarisation, topic analysis and document indexing. Considering the vast amount of text-based documents online in today's digital society, it is very useful to be able to extract keywords from online documents automatically to support large-scale textual analysis. Therefore, for many years the research community has been investigating automatic keyword extraction (AKE) methods, especially with the recent advancements in artificial intelligence (AI) and NLP. Despite these efforts, however, AKE has been shown to be a challenging task and AKE methods with very high performance are still to be found~\citep{papagiannopoulou2020}. Two main challenges are the lack of a precise definition of the AKE task and the lack of consistent performance evaluation metrics and benchmarks~\citep{merrouni2020automatic}. Since there is no consensus on the definition and characteristics of a \emph{keyword}, KE datasets created by researchers have different characteristics. Examples include the minimum/average/maximum numbers of keywords, if absent keywords (human-labelled keywords that do not appear in the text) are allowed, and what part-of-speech (PoS) tags such as verbs are accepted as valid keywords. This makes performance evaluation and comparison of AKE methods more difficult.

Based on whether a labelled training set is used, AKE methods reported in the literature can be grouped into unsupervised and supervised methods. Unsupervised methods include statistical, graph-based, embedding-based and/or language model-based methods, while supervised ones use either traditional or deep machine learning models~\citep{papagiannopoulou2020}. Surprisingly, for most AKE methods, semantic information has not been considered or only insufficiently considered to align the returned keywords with the semantic context of the input document~\citep{firoozeh2020}.

In this work, to fill the above-mentioned gap on the lack of or insufficient use of semantic information in the state-of-the-art (SOTA) AKE methods, we propose a \emph{universal} performance improvement approach for \emph{any} AKE methods. This approach serves as a post-processor that can consider semantic information more explicitly, with the support of PoS-tagging. To start with, we conducted an analysis of human-created `gold standard' keywords in 17 KE datasets to better understand some relevant characteristics of such keywords. Particularly, this analysis focuses on PoS-tag patterns, $n$-gram sizes, and the possible consideration of semantic information by human labellers when extracting keywords.

Our proposed approach is demonstrated using the following three post-processing steps that can be freely combined: (1) keeping candidate keywords with a desired PoS-tag only; (2) matching candidate keywords with one or more context-specific thesauri containing more semantically relevant terms; and (3) prioritising candidate keywords that appear as a valid Wikipedia named entity. We applied different combinations of the above three post-processing steps to five SOTA AKE methods, YAKE!~\citep{campos2020}, KP-Miner~\citep{elbeltagy2009}, RaKUn~\citep{skrlj2019}, LexRank \citep{ushio-etal-2021-back}, and SIFRank+~\citep{sun2020}, and compared the performances of the original methods with those of the enhanced versions. The experimental results with the 17 KE datasets showed that our proposed post-processing steps helped improve the performances of all the five SOTA AKE methods both \emph{consistently} (up to 100\% in terms of improved cases) and \emph{significantly} (between 10.2\% and 53.8\%, with an average of 25.8\%, in terms of F1-score and across all five methods), particularly when all the three steps are combined. Our work validates the possibility of using easy-to-use post-processing steps to enhance the semantic awareness of AKE methods and to improve their performance in real-world applications, a fact that has not been reported before (to the best of our knowledge).

The rest of the paper is organised as follows. Section~\ref{sec:related} briefly surveys AKE methods in the literature. The analysis of the human-created keywords in 17 KE datasets is given in Section~\ref{sec:groundtruth}. In Section~\ref{sec:methodology}, we present the methodology of our study. Section~\ref{sec:evaluation} explains the experimental setup for evaluation as well as the results. Finally, the paper is concluded with some further discussions in Section~\ref{sec:discussion}, and an overall summary in Section~\ref{sec:conclusion}.

\section{Related Work}
\label{sec:related}

\subsection{Unsupervised AKE Methods}

Some unsupervised AKE methods have been proposed, including statistical, graph-based, embedding-based and language model-based methods~\citep{papagiannopoulou2020}. Statistical AKE methods rely on some selected statistical metrics, e.g., term frequency, relevance to context, and co-occurrences, for ranking candidate keywords. One of the most used metrics is TF-IDF~\citep{jones1972statistical}, which combines two aspects of a term: term frequency within the input article, and the inverse document frequency across several domains. One AKE method using TF-IDF is KP-Miner~\citep{elbeltagy2009}, which also considers other metrics such as word length and word position. A more recent method in this category is YAKE!~\citep{campos2020}. It leverages a range of statistical metrics, such as casing, word position, word frequency, word relatedness to context, and how often a term appears in different sentences. Finally, LexSpec~\citep{ushio-etal-2021-back} makes use of lexical specificity -- a statistical metric to select the most representative keywords from a given text based on the hypergeometric distribution.

Graph-based AKE methods consider candidate keywords as nodes in a directed graph, often with weighted edges reflecting the syntactic/semantic relatedness of different keywords. They leverage graph-based methods, such as PageRank~\citep{BRIN1998107}, for ranking the nodes of the graph in terms of their overall importance. The earliest AKE method in this category is TextRank~\citep{mihalcea2004textrank}. It uses an unweighted graph of candidate keywords after filtering the ones that are not a noun or an adjective, and uses PageRank for ranking the nodes. As an extension to TextRank, SingleRank~\citep{wan2008} adds edge weights to the graph, which reflect the number of co-occurrences of the candidate keywords represented by any pair of two connected nodes. Another graph-based AKE method is RAKE~\citep{rose2010}, which builds a word-word co-occurrence graph, and assigns a score for each candidate by using word frequency and word degree. A more recent graph-based AKE method is RaKUn~\citep{skrlj2019}, which introduces meta-vertices by aggregating similar vertices, and employs load centrality metrics for candidate ranking. Finally, LexRank~\citep{ushio-etal-2021-back} and TFIDFRank~\citep{ushio-etal-2021-back} are two different enhanced versions of SingleRank, which use lexical specificity and TF-IDF, respectively.

Embedding-based AKE methods utilise word representation techniques, such as Doc2Vec~\citep{mikolov2013efficient} and GloVe~\citep{pennington2014glove}. An example method in this category is EmbedRank~\citep{bennani2018simple}, which uses sentence embeddings, and ranks candidate keywords in terms of cosine similarity. A more recent method is SIFRank~\citep{sun2020}, which combines sentence embedding model SIF and autoregressive pre-trained language model ELMo, and it was upgraded to SIFRank+ by position-biased weight to improve its performance for long documents. Lastly, MDERank~\citep{zhang-etal-2022-mderank} considers the similarity between the embeddings of the source document and its masked version for candidate ranking.

Apart from the AKE methods mentioned above, there also exist a number of AKE methods based on other techniques. \citet{rabby2020teket} proposed TeKET, a domain- and language-independent AKE technique utilising a binary tree for extracting final keywords from candidate ones. As another example, \citet{liu2009clustering} introduced an AKE algorithm based on term clustering considering semantic relatedness to identify the exemplar terms. The identified exemplar terms are, then, used to extract keywords.

\subsection{Supervised AKE Methods}

Although unsupervised methods are preferred for AKE, supervised methods have also been proposed. One of the earliest methods is KEA~\citep{witten2005kea}, which calculates TF-IDF scores and the position of the first occurrence of each candidate, and employs the Naive Bayes learning algorithm to decide if a candidate should be selected. More recently, there has been a growing interest in using deep learning for AKE. For example, \citet{basaldella2018} proposed an AKE method based on Bi-LSTM, which is capable of exploiting the context of each candidate word. Another AKE method, TNT-KID~\citep{martinc2021}, leverages transformers and allows users to train their own language model on a domain-specific corpus. A third example is TANN~\citep{wang2018}, an AKE method based on a topic-based artificial neural network model. It aims to improve the performance of AKE by transferring knowledge from a resource-rich source domain to an unlabelled or insufficiently labelled target domain. Finally, \citet{bordoloi2020} proposed a supervised variant of TextRank, leveraging a statistical supervised weighting scheme for terms to employ both global and local weights during keyword extraction.

\subsection{PoS-Tagging and Semantics in AKE}
\label{subsec:semantic-awareness}

Many AKE methods have considered how to extract more semantically meaningful keywords. For this purpose, PoS-tagging has been used so that extracted keywords are restricted to a pre-defined set of PoS-tag patterns, e.g., noun phrases only~\citep{hulth2003improved, pay2016, zervanou2010uvt}. Some methods utilise external knowledge to provide useful contextual information for extracting more semantically sensible keywords. For instance, \citet{li2014improved} proposed a TextRank-based AKE method that benefits from domain knowledge by using author-assigned keywords of scientific publications, and \citet{gazendam2010} proposed to use semantic relations between thesaurus terms for ranking candidate keywords without a reference corpus. Thesaurus relations have also been combined with machine learning techniques to improve the performance of AKE methods~\citep{hulth2001automatic, medelyan2006}. More recently, \citet{sheoran2022} leveraged domain-specific ontologies for aspect assignment of candidate keywords extracted from opinionated texts so that the selected candidates cover a maximum number of aspects.

Some AKE methods also make use of Wikipedia, a useful source of semantic information. \citet{shi2008} utilised Wikipedia to extract semantic features of candidate keywords. Their method constructs a semantic graph connecting candidate keywords to document topics based on the hierarchical relations extracted from Wikipedia, and semantic feature weights are assigned to candidate keywords with a link analysis algorithm. WikiRank is another AKE method leveraging Wikipedia~\citep{yu2018improving}. It employs the TAGME annotator~\citep{ferragina2010} to link meaningful word sequences in the input document to concepts in Wikipedia and constructs a semantic graph. Then, it transforms the KE task to an optimisation problem on the graph and tries to obtain the optimal keyword set that has the best coverage of the identified concepts. Finally, several embedding-based AKE methods utilise Wikipedia for pre-training and/or fine-tuning their underlying embedding methods~\citep{bennani2018simple, papagiannapoulou2018}.

Compared with existing AKE methods that have considered PoS-tagging or semantic information more explicitly, our proposed approach is more universal and can be applied to \emph{any} AKE methods as a post-processor, which simply re-evaluate candidate keywords generated by an AKE method before top $n$ keywords are returned. Our approach is easily generalisable and can be used flexibly to eliminate candidate keywords that are unlikely to be a keyword and prioritise those that are more likely to be a keyword.

\section{Analysis of Human-Created Keywords}
\label{sec:groundtruth}

Our proposed approach was motivated by some of our observations regarding how human labellers extracted ``golden'' (i.e., ground truth) keywords in 17 KE datasets. Such observations also helped us to determine some specific details such as the parameters used in our proposed approach. In the following, we describe the 17 datasets we used and the key observations.

\subsection{Datasets Inspected}
\label{subsec:datasets}

Considering the subjectivity of the keyword extraction task, a standard approach has not been established to follow for constructing keyword extraction datasets~\citep{zesch2009approximate}. This brings an extreme diversity to the datasets constructed so far, which makes comprehensive tests of keyword extraction algorithms harder. Therefore, to achieve a better understanding of human-created keywords, we aimed at collecting a wide range of representative KE datasets used in the literature. With this respect, research papers corresponding to SOTA AKE methods and relevant surveys have been searched on multiple research databases, including Google Scholar and Scopus, with the keywords ``automatic keyword extraction'' and ``automatic keyphrase extraction''. Then, the collected papers were reviewed to identify datasets used by other researchers, and the publicly available datasets were downloaded. Besides, multiple collections of AKE datasets have been found through different GitHub repositories\footnote{Examples include \url{https://github.com/LIAAD/KeywordExtractor-Datasets}, \url{https://github.com/boudinfl/ake-datasets}, and \url{https://github.com/SDuari/Keyword-Extraction-Datasets}.}. In total, we were able to collect 17 datasets covering multiple contexts, including agriculture, computer science and health, and several types of documents, such as scientific papers, news, theses and abstracts. However, we excluded datasets containing short-text documents, such as tweets, since they tend to contain fewer candidate keywords, which could negatively impact the informativeness of our analysis. In addition, we only selected English datasets to limit the scope of our study to the English language due to the lack of a sufficient amount of non-English datasets and English being the only language shared by the authors of this paper. Further details regarding the datasets can be seen in Table~\ref{tab:datasets}.

\begin{table}[!htb]
\scriptsize
\tabcolsep=0.05cm
\centering
\caption{Basic information about the 17 datasets}
\label{tab:datasets}
\begin{tabularx}{\linewidth}{ccccccc}
\toprule
\textbf{Dataset} & \textbf{Content} & \textbf{Context} & \textbf{Size} & \textbf{Avg.~\#(Keys)} & \textbf{Abs.~Keys} & \textbf{Annotators}\tnote{1}\\
\midrule
KPCrowd~\citep{marujo2013supervised} & News & Misc. & 500 & 48.92 & 13.5\% & Readers\\
citeulike180~\citep{medelyan2009human} & Paper & Misc. & 183 & 18.42 & 32.2\% & Readers\\
DUC-2001~\citep{wan2008} & News & Misc. & 308 & 8.1 & 3.7\% & Readers\\
fao30~\citep{medelyan2008domain} & Paper & Agr. & 30 & 33.23 & 41.7\% & Experts\\
fao780~\citep{medelyan2008domain} & Paper & Agr. & 779 & 8.97 & 36.1\% & Experts\\
Inspec~\citep{hulth2003improved} & Abstract & CS & 2,000 & 14.62 & 37.7\% & Experts\\
KDD~\citep{gollapalli2014extracting} & Abstract & CS & 755 & 5.07 & 53.2\% & Authors\\
KPTimes (test)~\citep{gallina2019kptimes} & News & Misc. & 20,000 & 5.0 & 54.7\% & Editors\\
Krapivin2009~\citep{krapivin2009large} & Paper & CS & 2,304 & 6.34 & 15.3\% & Authors\\
Nguyen2007~\citep{nguyen2007keyphrase} & Paper & CS & 209 & 11.33 & 17.8\% & Authors \& Readers\\
PubMed~\citep{gay2005semi} & Paper & Health & 500 & 15.24 & 60.2\% & Authors\\
Schutz2008~\citep{schutz2008keyphrase} & Paper & Health & 1,231 & 44.69 & 13.6\% & Authors\\
SemEval2010~\citep{kim2010semeval} & Paper & CS & 243 & 16.47 & 11.3\% & Authors \& Readers\\
SemEval2017~\citep{augenstein2017semeval} & Paragr & Misc. & 493 & 18.19 & 0.0\% & Experts \& Readers\\
theses100\tnote{2} & Thesis & Misc. & 100 & 7.67 & 47.6\% & Unknown\\
wiki20~\citep{medelyan2008topic} & Report & CS & 20 & 36.50 & 51.2\% & Readers\\
WWW~\citep{gollapalli2014extracting} & Abstracts & CS & 1,330 & 5.80 & 55.0\% & Authors\\
\bottomrule
\end{tabularx}
\begin{tablenotes}
\item[1] Experts: Professional indexers assigned for annotation, Readers: People recruited for annotation regardless of their expertise, Authors: The authors of the document annotated
\item[2] \url{https://github.com/LIAAD/KeywordExtractor-Datasets#theses100}
\end{tablenotes}
\end{table}

\subsection{Observations: PoS-Tag Patterns}
\label{subsec:pos-tag-patterns}

There has been a lot of research on linguistic properties of different multi-word expression types, such as collocations~\citep{smadja1993} and technical terms~\citep{justeson_katz_1995}. In addition, various PoS-tag patterns have been proposed in the literature to identify noun phrases, which have been commonly considered a major indicator of keyword candidates~\citep{ajallouda2022}. However, these are unable to properly explain the linguistic properties of \emph{keywords} used in AKE research because of the lack of linguistic standards for human-created keywords. Therefore, firstly, we reviewed the structure of human-created keywords in the 17 datasets, in terms of the used PoS-tag patterns. For this purpose, we used the NLTK~\citep{bird2006nltk} library's PoS tagger and computed the distribution of different PoS-tag patterns. As shown in Table~\ref{tab:pos-patterns}, nine of the top ten PoS-tag patterns correspond to either noun or gerund phrases. The only non-noun/gerund pattern in the top ten PoS-tag patterns is a single adjective (JJ), with an average percentage of 6.85\%. The top ten PoS-tag patterns count 80\% of all patterns. These observations imply that leveraging knowledge about how human labellers define keywords based on PoS-tag patterns for a specific domain can potentially help improve the performance of any AKE methods for the corresponding domain.

\begin{table}[!htb]
\caption{Percentages of top 10 PoS-tag patterns across 17 datasets. PoS tags: NN -- noun (singular), NNS -- noun (plural), JJ -- adjective, VBG -- verb gerund.}
\scriptsize
\centering
\tabcolsep=0.02cm
\renewcommand{\arraystretch}{1.3}
\begin{tabular}{*{11}{|c}|}
\hline
\textbf{Dataset} & NN & NN NN & JJ NN & NNS & JJ & JJ NNS & NN NNS & JJ NN NN & VBG & NN NN NN\\
\hline
KPCrowd & 31.38 & 2.18 & 3.29 & 11.65 & 10.13 & 0.95 & 0.95 & 0.26 & 5.27 & 0.17\\
citeulike180 & 48.71 & 7.03 & 4.78 & 12.93 & 12.74 & 1.61 & 1.56 & 0.15 & 1.95 & 0.05\\
DUC-2001 & 19.13 & 15.90 & 15.28 & 10.49 & 1.80 & 8.73 & 10.16 & 3.65 & 0.28 & 1.52\\
fao30 & 32.60 & 14.68 & 7.92 & 15.84 & 5.06 & 6.62 & 9.35 & 0.00 & 0.78 & 0.26\\
fao780 & 29.56 & 14.11 & 9.11 & 15.18 & 3.78 & 6.02 & 10.88 & 0.06 & 1.21 & 0.04\\
Inspec & 19.05 & 12.57 & 12.49 & 6.64 & 3.85 & 8.11 & 5.95 & 4.35 & 1.11 & 2.50\\
KDD & 27.93 & 13.49 & 9.06 & 5.89 & 9.25 & 5.13 & 3.55 & 2.22 & 4.81 & 0.76\\
KPTimes & 15.32 & 16.65 & 15.67 & 4.27 & 2.83 & 8.62 & 6.26 & 2.92 & 1.76 & 1.51\\
Krapivin2009 & 35.15 & 4.70 & 4.06 & 14.14 & 5.67 & 2.17 & 1.47 & 0.27 & 0.95 & 0.17\\
Nguyen2007 & 20.85 & 19.83 & 11.31 & 4.84 & 2.53 & 4.79 & 3.37 & 3.06 & 1.51 & 2.66\\
PubMed & 30.88 & 9.23 & 3.87 & 15.43 & 12.01 & 3.51 & 5.50 & 0.77 & 0.56 & 2.03\\
Schutz2008 & 30.15 & 6.20 & 10.61 & 18.63 & 10.91 & 5.04 & 3.19 & 1.61 & 0.31 & 0.66\\
SemEval2010 & 19.45 & 21.74 & 21.54 & 0.08 & 3.20 & 0.17 & 0.06 & 6.40 & 0.42 & 3.15\\
SemEval2017 & 14.57 & 8.73 & 9.00 & 7.23 & 2.12 & 5.95 & 4.46 & 3.31 & 0.66 & 1.62\\
theses100 & 27.88 & 8.55 & 5.39 & 9.48 & 15.24 & 6.13 & 4.28 & 0.00 & 1.30 & 0.19\\
wiki20 & 41.91 & 18.65 & 11.06 & 1.49 & 6.60 & 0.50 & 1.82 & 2.81 & 2.81 & 0.99\\
WWW & 32.33 & 13.44 & 8.98 & 5.41 & 8.74 & 3.88 & 3.88 & 1.63 & 2.86 & 1.05\\
\hline
\textit{Average (\%)} & \textit{28.05} & \textit{12.22} & \textit{9.61} & \textit{9.39} & \textit{6.85} & \textit{4.59} & \textit{4.51} & \textit{1.97} & \textit{1.68} & \textit{1.13}\\
\hline
\end{tabular}
\label{tab:pos-patterns}
\end{table}

\subsection{Observations: $n$-Gram Size}
\label{subsec:observations_ngram-size}

AKE methods generally include a parameter for the maximum $n$-gram size, corresponding to the maximum number of words a keyword is allowed to contain. Although it is well-known that multi-word expressions (MWEs) are more likely to be of length two to three in English~\citep{choueka1988}, it is less clear how human labellers of the 17 datasets were instructed to consider the $n$-gram size. Therefore, we analysed the golden keywords across the 17 datasets to see how human labellers decided on the $n$-gram sizes. On average, bigrams ($n=2$) constitute 45.55\% of the golden keywords in the 17 datasets, while this rate is 36.45\% for unigrams ($n=1$) and 12.73\% for trigrams ($n=3$). In addition, the percentages for keywords with $n \geq 4$ are considerably low -- 5.12\% on average. More detailed statistics can be seen in Table~\ref{tab:ngram}. These results show that human labellers largely used two or three as the maximum $n$-gram size, covering 82.01\% and 94.74\% of the golden keywords across the different datasets, respectively. The results are aligned with those in the research literature on MWEs. Based on such observations, we can see that AKE methods could benefit from focusing more on keywords with a shorter word length.

\begin{table}[!htb]
\scriptsize
\renewcommand{\arraystretch}{1.3}
\centering
\caption{$n$-gram distributions of the 17 datasets}
\label{tab:ngram}
\begin{tabular}{ | c | cccc | c|c |}
\hline
\textbf{Dataset} & $n=1$ & $n=2$ & $n=3$ & \textbf{$n\geq 4$} & $n=1,2$ & $1\leq n\leq 3$\\ 
\hline
KPCrowd & \textbf{73.78} & 18.47 & 4.90 & 2.83 & 92.25 & 97.15\\
citeulike180 & \textbf{77.10} & 19.98 & 2.79 & 0.09 & 97.08 & 99.87\\
DUC-2001 & 17.32 & \textbf{61.29} & 17.73 & 3.66 & 78.61 & 96.34\\
fao30 & 43.02 & \textbf{52.74} & 3.41 & 0.83 & 95.76 & 99.17\\
fao780 & 42.32 & \textbf{53.72} & 3.62 & 0.34 & 96.04 & 99.66\\
Inspec & 16.44 & \textbf{53.68} & 23.05 & 6.84 & 70.12 & 93.17\\ 
KDD & 25.48 & \textbf{56.32} & 13.97 & 4.24 & 81.80 & 95.77\\
KPTimes & \textbf{46.68} & 34.39 & 12.55 & 6.38 & 81.07 & 93.62\\
Krapivin2009 & 18.95 & \textbf{61.61} & 15.74 & 3.70 & 80.56 & 96.30\\
Nguyen2007 & 27.53 & \textbf{49.96} & 15.42 & 6.97 & 77.49 & 92.91\\
PubMed & 35.79 & \textbf{43.74} & 15.90 & 4.58 & 79.53 & 95.43\\
Schutz2008 & \textbf{57.83} & 30.22 & 8.15 & 1.67 & 88.05 & 96.20\\
SemEval2010 & 20.05 & \textbf{52.97} & 20.66 & 6.31 & 73.02 & 93.68\\
SemEval2017 & 25.23 & \textbf{33.74} & 17.19 & 23.84 & 58.97 & 76.16\\
theses100 & 31.63 & \textbf{50.37} & 11.09 & 6.90 & 82.00 & 93.09\\
wiki20 & 26.20 & \textbf{53.52} & 18.17 & 2.11 & 79.72 & 97.89\\
WWW & 34.36 & \textbf{47.71} & 12.15 & 5.78 & 82.07 & 94.22\\
\hline
\textit{Average (\%)} & \textit{36.45} & \textit{\textbf{45.55}} & \textit{12.73} & \textit{5.12} & \textit{82.01} & \textit{94.74}\\
\hline
\end{tabular}
\end{table}

\subsection{Observations: Semantic Information}
\label{subsec:observations_semantic}

Finally, we analysed the human-created keywords to see if human labellers explicitly or implicitly relied on semantic information to select keywords. We first calculated the percentage of golden keywords that are covered by Wikipedia across all the datasets. This quantitative analysis indicated that, on average, 64.39\% of the golden keywords are Wikipedia named entities, i.e., titles of Wikipedia articles. This interesting (previously unreported) finding justifies that Wikipedia can be a very useful knowledge base for AKE algorithms as it covers so many golden keywords chosen by human labellers for all the 17 datasets we chose. Although unexpected, this finding can be explained by the diversity and richness of the content of Wikipedia. More detailed results of the analysis can be seen in Table~\ref{tab:semantic-wiki}.

\begin{table}[!htb]
\scriptsize
\caption{The percentages of golden keywords covered by Wikipedia.}
\centering
\begin{tabular}{cccc}
\toprule
\textbf{Dataset} & \textbf{\%} & \textbf{Dataset} & \textbf{\%} \\ 
\midrule
KPCrowd & 71.77 & Nguyen2007 & 52.19\\
citeulike180 & 83.78 & PubMed & 81.28\\
DUC-2001 & 51.05 & Schutz2008 & 67.43\\
fao30 & 80.97 & SemEval2010 & 41.27\\
fao780 & 79.00 & SemEval2017 & 31.02\\
Inspec & 39.08 & theses100 & 68.82\\
KDD & 62.92 & wiki20 & 89.01\\
KPTimes & 79.09 & WWW & 63.83\\
Krapivin2009 & 52.12\\
\bottomrule
\end{tabular}
\label{tab:semantic-wiki}
\end{table}

In addition to Wikipedia named entities, we also manually inspected many golden keywords and observed that many collected datasets contain domain-specific golden keywords. This observation indicates that considering domain-specific terms can potentially help improve the performance of AKE methods, too.

\section{Methodology}
\label{sec:methodology}

\subsection{Problem Definition and Our Proposed Approach}

Suppose $W^C(D) = \{w^C_i\}_{i=1}^m$ denotes $m$ candidate keywords generated from a document $D$ by an AKE method. In addition, let $W^S(D) \subset W^C(D)$ denotes $n \leq m$ keywords produced by the AKE method. Finally, let $W(D) = \{w_i\}_{i=1}^t$ be the set of ground truth keywords an ideal AKE method should extract from $D$. Given the above notations, our goal is to find post-processing methods that can minimise $|W(D) - W^S(D)|$ (false negatives) and $|W^S(D) - W(D)|$ (false positives). Among the two types of errors, reducing false negatives is more important than reducing false positives, but given the fact that $n$ cannot be too large to make the results manageable balancing both types of errors is still very important. Typically, AKE methods select keywords by assigning a numerical score $s_i$ to each candidate keyword $w_i$, and then return the top $n$ keywords with the highest\footnote{Although some AKE methods, e.g., YAKE!, use smaller scores for better keywords, here, for the sake of simplicity, we assume that a higher score means a more preferred keyword.} scores. Our proposed approach can work with any AKE method with such a scoring system, and it aims to re-adjust such scores so that true positive keywords' scores will more likely increase and true negative keywords' scores will more likely decrease.

\begin{figure}[htb!]
\includegraphics[width=\textwidth]{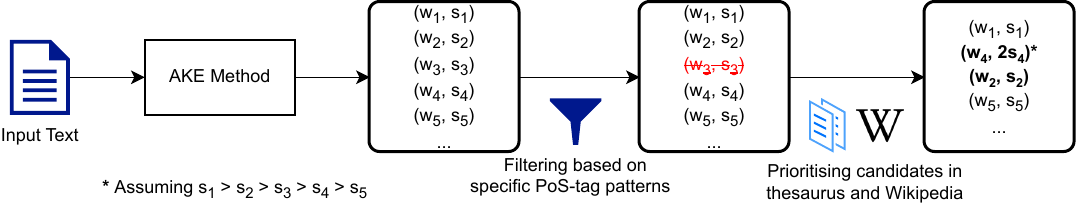}
\caption{The overview of the proposed post-processing approach.}
\label{fig:overview}
\end{figure}

Informed by the findings presented in Chapter~\ref{sec:groundtruth}, our proposed approach is based on three general post-processing steps that can be applied to any baseline AKE methods as shown in Figure~\ref{fig:overview}: 1) removing candidate keywords with an unlikely PoS-tag pattern by zeroing its score ($s_i=0$), 2) using one or more context-aware (i.e., domain-specific) thesauri to prioritise important candidate keywords for the target domain ($s_i=c_is_i$, where $c_i$ is an amplifying factor larger than 1), and 3) prioritising candidate keywords that are Wikipedia named entities ($s_i=w_is_i$, where $w_i$ is another amplifying factor larger than 1). Note that the amplifying factor $c_i$ and $w_i$ can be a static value for all prioritised keywords (so independent of $i$) or a keyword-dependent factor, depending on an importance score of each candidate keyword in the thesauri and Wikipedia, e.g., $c_i$ can be proportional to the word frequency in the thesauri and $w_i$ can be proportional to the size of the Wikipedia entry or the number of references to the entry.

In the following subsections, we explain the three steps in more detail.

\subsection{Filtering Specific PoS-Tag Patterns}
\label{subsec:filtering-specific-patterns}

As mentioned in Section~\ref{subsec:semantic-awareness}, PoS-tagging has been extensively used in AKE methods to consider morpho-syntactic features. Motivated by the observations in Section~\ref{subsec:pos-tag-patterns}, we attempted to leverage a PoS-tagger to filter out candidate keywords labelled with unlikely PoS-tag patterns. More precisely, candidate keywords that do not conform with any of the following PoS-tag patterns were discarded: (i) \emph{simple nouns and noun phrases} -- one or more nouns/gerunds (optionally with one or more adjectives appearing before the first noun); (ii) two or more \emph{simple nouns and/or noun phrases} connected by one or more prepositions or conjunctions\footnote{Examples include ``quality of service" and ``buyer and seller" from the SemEval2010 dataset. Although none of the possible PoS-tag patterns conforming to this criterion are among the most common patterns presented in Table~\ref{tab:pos-patterns} individually, they collectively constitute 1.3\% of all patterns across the 17 datasets.}; and (iii) a single adjective.

In the PoS-tag patterns mentioned above, \emph{nouns} and \emph{adjectives} mean any PoS tags that can provide the corresponding functionality in a sentence. Therefore, nouns also include gerunds, and adjectives also include past participle verbs. Considering the most common PoS-tag patterns mentioned in Section~\ref{subsec:pos-tag-patterns}, our proposed PoS-tag patterns correspond to over 90\% of the patterns observed across all the 17 datasets. We used NLTK to extract PoS tags for each term in the input documents. For pattern matching, we took the advantage of regular expressions. Since regular expressions return the longest possible matches, we extracted the shorter matches from the longest ones separately.

Note that the proposed PoS-tag patterns can be further changed to reflect any domain-specific needs, e.g., we observed that gerunds are quite uncommon in the health domain so they can be removed if preferred.

\subsection{Context-Aware Thesauri}
\label{subsec:Method-Context-AwareThesauri}

Context means any kind of domain, topic or field that has its own set of terms semantically specific to itself. While the set of terms specific to a context can be covered in a more structured vocabulary, such as a thesaurus or an ontology, a simple word list can often be sufficient for the purpose of AKE. As reported in Section~\ref{subsec:observations_semantic}, many keywords are related to the context of the input text, contextual consideration can be quite useful for AKE. Therefore, we propose to make use of external resources to inform AKE methods more about semantically useful keywords for the relevant domain. More specifically, we proposed to integrate one or more domain-specific thesauri, which contain terms specific to a target context, and to prioritise candidate keywords included in such thesauri. At the implementation level, we introduce a weight for each candidate keyword and increase the weight of any candidate keyword appearing in one of the thesauri. In our experiments, we doubled the weights of candidate keywords in a thesaurus. However, the actual weight increase can be a parameter, which can be empirically determined based on some training data or qualitative evidence observed. To determine if a candidate keyword exists in a given thesaurus, we applied exact matching with lemmatisation. Although using stemming with exact matching is a more common practice in AKE~\citep{papagiannopoulou2020}, we preferred to use lemmatisation due to its context-awareness. In our experiments, we focused on thesauri with a single context, but using multiple contexts in a single thesaurus is of course also possible. Regarding integrating relevant thesauri, we considered two different approaches explained below.

\paragraph{Manual Context Consideration:} This approach is more useful when documents processed by an AKE method are known to belong to a specific context. It utilises one or more thesauri containing a list of terms relevant to the context, which are given a higher weight for prioritisation by the AKE method. In our experiments, we assigned a single domain-specific thesaurus to each of the datasets to represent the relevant context. Note that it is possible that multiple contexts and multiple thesauri are used in some applications of AKE.

\paragraph{Automatic Context Identification:} Considering the wide range of applications in which AKE methods can be utilised, manually providing a thesaurus for each input document may not be very usable. Therefore, we also studied how to identify the context of an input document automatically, which can allow assigning a different context and a corresponding thesaurus automatically. This can be achieved by building a machine learning-based classifier, which produces a class label representing the context or a context-specific thesaurus of a given document or its abstract. 

Once the classifier predicts the context of an input abstract, we identify a thesaurus corresponding to the context, as defined in a context-to-thesaurus look-up table, to inform the AKE method. Unlike the manual approach, automatic identification allows us to use a different thesaurus for each document in the dataset, therefore can be applied to many real-world scenarios where the documents processed can belong to multiple contexts.

\subsection{Wikipedia Named Entities}

Based on the Wikipedia-related observations reported in Section~\ref{subsec:observations_semantic}, we propose to use Wikipedia as a \emph{context-independent} thesaurus to improve the performance of any AKE methods working in any context(s). Similar to how a thesaurus can be used, we prioritise the candidate keywords covered by Wikipedia as an entry by increasing their weight. Then, we apply exact matching with lemmatisation to identify if a candidate keyword is a Wikipedia named entity. Since Wikipedia also contains a vast amount of entries with too general semantic meanings, e.g., unigrams such as `father', `school' and `table' that are normally already well covered by most AKE methods, we utilised the NLTK's \emph{words} corpus (i.e., a wordlist including common English dictionary words) to identify such unigrams and remove them from the Wikipedia entities that will be prioritised in our post-processing step. For the Wikipedia named entities, we used the 2021-10-01 version of the English Wikipedia dump\footnote{\url{https://archive.org/download/enwiki-20211001}}, containing only page titles. We firstly cleaned the dump data by removing the disambiguation tags\footnote{\url{https://en.wikipedia.org/wiki/Wikipedia:Disambiguation\#Naming_the_disambiguation_page}} added next to the title by Wikipedia. Then, we normalised the data with lemmatisation and lower-casing by following the common practice.

\section{Experiments and Results}
\label{sec:evaluation}

\subsection{Evaluation Metrics}

As the evaluation metrics, we used precision, recall and F1-score at the top ten keywords, which have been commonly used in AKE evaluation~\citep{papagiannopoulou2020}. Furthermore, we adopted micro-averaging and exact matching with stemming when calculating the scores.

\subsection{Selecting Baseline Methods}

To show the effectiveness and generalisability of the proposed methods, we first attempted to identify some representative AKE algorithms with different key characteristics for our experiments. We reviewed existing AKE algorithms in terms of multiple aspects, e.g., recency, ease of reconfiguration, and if they already use one or more of our proposed methods by any means, as shown in Table~\ref{tab:ake-methods}. These methods are considered more representative because they have open-source implementations, are applicable to any document type, were validated on a number of datasets, and do not require training (i.e., unsupervised so that it is easier to use and less likely to have generalisation problems)\footnote{Unsupervised AKE methods have become more popular for this reason. Most implementations of supervised methods are also harder to reconfigure.}. Among these methods, we selected two statistical methods, i.e., KP-Miner and YAKE!, two graph-based methods, i.e., RaKUn and LexRank, and an embedding-based method, i.e., SIFRank+, as baseline methods for our experiments. Since SIFRank+ is very computationally costly, we used only seven of the datasets, containing shorter documents (i.e., KPCrowd, DUC-2001, Inspec, KDD, KPTimes, SemEval2017 and WWW) for its evaluation when our methods were applied.

For the implementations of the selected AKE methods, we utilised the PKE~\citep{b2016} library for KP-Miner and the original implementations of the other four. We used the default parameters for all the methods, except the maximum $n$-gram size parameter. Considering the $n$-gram size across the datasets being mostly limited up to 3, as mentioned in Section~\ref{subsec:observations_ngram-size}, we set the maximum $n$-gram size to be 3.

\begin{table}[!htb]
\scriptsize
\centering
\caption{An overview of some existing open-source unsupervised AKE methods, showing a number of key characteristics.}
\begin{tabular}{l *{4}{c}}
\toprule
\textbf{Method} & \textbf{Easy to} & \textbf{PoS-tagging} & \textbf{Thesaurus} & \textbf{Wikipedia}\\ 
& \textbf{Reconfigure} & & &\\
\midrule
\textit{Statistical Methods}\\
KP-Miner~\citep{elbeltagy2009} & \checkmark & -- & -- & --\\
YAKE!~\citep{campos2020} & \checkmark & -- & -- & --\\
LexSpec~\citep{ushio-etal-2021-back} & \checkmark & \checkmark & -- & --\\
\midrule
\textit{Graph-based Methods}\\
TextRank~\citep{mihalcea2004textrank} & \checkmark & \checkmark & -- & --\\
SingleRank~\citep{wan2008} & \checkmark & \checkmark & -- & --\\
RAKE~\citep{rose2010} & \checkmark & -- & -- & --\\
RaKUn~\citep{skrlj2019} & \checkmark & -- & -- & --\\
LexRank~\citep{ushio-etal-2021-back} & \checkmark & \checkmark & -- & --\\
TFIDFRank~\citep{ushio-etal-2021-back} & \checkmark & \checkmark & -- & --\\
\midrule
\textit{Embeddings-based Methods}\\
EmbedRank~\citep{bennani2018simple} & \checkmark & \checkmark & -- & \checkmark\\
SIFRank~\citep{sun2020} & \checkmark & \checkmark & -- & \checkmark\\
SIFRank+~\citep{sun2020} & \checkmark & \checkmark & -- & \checkmark\\
MDERank~\citep{zhang-etal-2022-mderank} & -- & \checkmark & -- & \checkmark\\
\bottomrule
\end{tabular}
\label{tab:ake-methods}
\end{table}

\subsection{PoS-Tag Patterns}
\label{subsec:pos-tagging}

As the first step, we applied our PoS-tagging-based post-processing approach to the selected AKE methods and evaluated on all the datasets. Results show that the proposed approach improved all the methods except SIFRank+ on average, in terms of precision, recall, and F1-score. While KP-Miner achieved better performance for 14 of the 17 datasets with an average of 6.08\% in F1-score, RaKUn was improved by a cross-dataset average of 4.46\% for 14 of the 17 datasets. We observed the most change in the performance of YAKE!\ -- it was improved in 16 of the 17 datasets by a cross-dataset average of 18.05\%. We believe this is because YAKE!\ does not benefit from linguistic features as a more language-independent (multilingual) approach. Finally, we observed a limited improvement in the scores of LexRank (0.84\% on average) for 12 of the 17 datasets, and a slight decrease in the performance of SIFRank+, which is likely due to the fact that these two methods already use PoS-tagging-based filtering. The obtained scores for YAKE!\ and SIFRank+ are shown in Tables~\ref{tab:yake-pos} and \ref{tab:sifrank-pos}, respectively, as examples. These results provide new evidence for the effectiveness of PoS-tagging in AKE algorithms and imply that there is still room to improve the use of PoS-tagging in many AKE methods.

\begin{table}[!htb]
\caption{Comparison of the precision, recall, and F1-score of the original YAKE!\ and the one utilising PoS-tagging, at 10 extracted keywords}
\scriptsize
\centering
\begin{tabular}{*{7}{c}}
\toprule
\multirow{2}{*}{\textbf{Dataset}} & \multicolumn{3}{c}{\textbf{YAKE!}} & \multicolumn{3}{c}{\textbf{YAKE!+PoS}}\\
\cmidrule(lr){2-4} \cmidrule(lr){5-7}
& \textbf{P\%} & \textbf{R\%} & \textbf{F1\%} & \textbf{P\%} & \textbf{R\%} & \textbf{F1\%}\\
\midrule
KPCrowd & 24.20 & 4.92 & 8.17 & \textbf{33.98} & \textbf{6.90} & \textbf{11.47}\\
citeulike180 & 23.11 & 13.27 & 16.86 & \textbf{25.68} & \textbf{14.74} & \textbf{18.73}\\
DUC-2001 & 12.01 & 14.87 & 13.29 & \textbf{17.44} & \textbf{21.58} & \textbf{19.29}\\
fao30 & 22.00 & 6.83 & 10.42 & \textbf{25.33} & \textbf{7.86} & \textbf{12.00}\\
fao780 & 11.93 & 14.95 & 13.27 & \textbf{13.18} & \textbf{16.52} & \textbf{14.67}\\
Inspec & 19.82 & 14.05 & 16.44 & \textbf{24.57} & \textbf{17.41} & \textbf{20.38}\\
KDD & \textbf{6.01} & \textbf{14.68} & \textbf{8.53} & 5.83 & 14.23 & 8.27\\
KPTimes & 7.97 & 15.83 & 10.61 & \textbf{11.37} & \textbf{22.58} & \textbf{15.12}\\
Krapivin2009 & 9.54 & 17.88 & 12.44 & \textbf{9.93} & \textbf{18.61} & \textbf{12.95}\\
Nguyen2007 & 19.00 & 15.82 & 17.26 & \textbf{19.19} & \textbf{15.98} & \textbf{17.43}\\
PubMed & 7.28 & 5.11 & 6.01 & \textbf{8.66} & \textbf{6.08} & \textbf{7.15}\\
Schutz2008 & 37.29 & 8.06 & 13.26 & \textbf{47.63} & \textbf{10.30} & \textbf{16.93}\\
SemEval2010 & 20.37 & 13.08 & 15.93 & \textbf{20.82} & \textbf{13.37} & \textbf{16.28}\\
SemEval2017 & 20.61 & 11.91 & 15.10 & \textbf{29.41} & \textbf{17.00} & \textbf{21.55}\\
theses100 & 9.40 & 14.09 & 11.28 & \textbf{10.50} & \textbf{15.74} & \textbf{12.60}\\
wiki20 & 19.50 & 5.49 & 8.57 & \textbf{22.00} & \textbf{6.20} & \textbf{9.67}\\
WWW & 6.49 & 13.47 & 8.76 & \textbf{6.58} & \textbf{13.66} & \textbf{8.88} \\
\midrule
\textbf{\textit{Avg.~Score (\%)}} & 16.27 & 12.02 & 12.13 & \textbf{19.54} & \textbf{14.04} & \textbf{14.32}\\
\textbf{\textit{Improvement (\%)}} & & & & \textcolor{olive}{\textit{20.10}} & \textcolor{olive}{\textit{16.81}} & \textcolor{olive}{\textit{18.05}}\\
\bottomrule
\end{tabular}
\label{tab:yake-pos}
\end{table}

\begin{table}[!htb]
\caption{Comparison of the precision, recall, and F1-score of the original SIFRank+ and the one utilising PoS-tagging, at 10 extracted keywords}
\scriptsize
\centering
\begin{tabular}{*{7}{c}}
\toprule
\multirow{2}{*}{\textbf{Dataset}} & \multicolumn{3}{c}{\textbf{SIFRank+}} & \multicolumn{3}{c}{\textbf{SIFRank+ + PoS}} \\ \cmidrule(lr){2-4} \cmidrule(lr){5-7}
 & \textbf{P\%} & \textbf{R\%} & \textbf{F1\%} & \textbf{P\%} & \textbf{R\%} & \textbf{F1\%} \\
\midrule
KPCrowd & 26.08 & 5.30 & 8.81 & \textbf{26.20} & \textbf{5.32} & \textbf{8.85} \\
DUC-2001 & \textbf{28.34} & \textbf{35.09} & \textbf{31.36} & 27.86 & 34.49 & 30.82 \\
Inspec & \textbf{35.68} & \textbf{25.29} & \textbf{29.60} & 35.10 & 24.88 & 29.12 \\
KDD & \textbf{5.68} & \textbf{13.87} & \textbf{8.06} & 4.42 & 10.80 & 6.28 \\
KPTimes & \textbf{7.92} & \textbf{15.74} & \textbf{10.54} & 7.74 & 15.37 & 10.30 \\
SemEval2017 & \textbf{41.66} & \textbf{24.08} & \textbf{30.52} & 40.16 & 23.21 & 29.42 \\
WWW & \textbf{6.59} & \textbf{13.69} & \textbf{8.90} & 5.26 & 10.93 & 7.10 \\ \midrule
\textbf{\textit{Avg.~Score (\%)}} & \textbf{21.71} & \textbf{19.01} & \textbf{18.26} & 20.96 & 17.86 & 17.41 \\
\textbf{\textit{Improvement (\%)}} & & & & \textcolor{red}{\textit{-3.45}} & \textcolor{red}{\textit{-6.05}} & \textcolor{red}{\textit{-4.65}} \\
\bottomrule
\end{tabular}
\label{tab:sifrank-pos}
\end{table}

Finally, we studied how tailoring the selected PoS-tag patterns according to domain-specific needs may affect the performance of AKE methods. To this end, we considered the example given in Section~\ref{subsec:filtering-specific-patterns}, i.e., the observation that gerunds are rarely seen as a keyword in the health domain. We selected the health datasets (i.e., PubMed and Schutz2008) from our collection and applied the tailored PoS-tag-based filtering that disregards gerunds. For this experiment, we used YAKE!\ since it is more sensitive to linguistic-based improvements as a language-independent algorithm. As shown in Table~\ref{tab:yake-pos-health}, the tailored filtering approach provided some small improvements to our original filtering proposal in terms of precision, recall, and F1-score. The limited improvement is likely due to the small percentage of gerunds as candidate keywords.

\begin{table}[!htb]
\caption{Comparison of the precision, recall, and F1-score of YAKE!\ when the original (PoS) and the tailored (PoS*) filtering approaches are used, at 10 extracted keywords}
\scriptsize
\centering
\begin{tabular}{*{7}{c}}
\toprule
\multirow{2}{*}{\textbf{Dataset}} & \multicolumn{3}{c}{\textbf{YAKE!+PoS}} & \multicolumn{3}{c}{\textbf{YAKE!+PoS*}}\\ \cmidrule(lr){2-4} \cmidrule(lr){5-7}
& \textbf{P\%} & \textbf{R\%} & \textbf{F1\%} & \textbf{P\%} & \textbf{R\%} & \textbf{F1\%}\\
\midrule
PubMed & 8.66 & 6.08 & 7.15 & \textbf{8.70} & \textbf{6.11} & \textbf{7.18}\\
Schutz2008 & 47.63 & 10.30 & 16.93 & \textbf{47.80} & \textbf{10.34} & \textbf{17.00}\\
\midrule
\textbf{\textit{Avg.~Score (\%)}} & 28.15 & 8.19 & 12.04 & \textbf{28.25} & \textbf{8.23} & \textbf{12.09}\\
\textbf{\textit{Improvement (\%)}} & & & & \textcolor{olive}{\textit{0.36}} & \textcolor{olive}{\textit{0.49}} & \textcolor{olive}{\textit{0.42}}\\
\bottomrule
\end{tabular}
\label{tab:yake-pos-health}
\end{table}

\subsection{Context-Aware Thesauri}

For this step, we selected 10 datasets mentioned in Section~\ref{subsec:datasets} that have a particular context. The included contexts (and datasets) are agriculture (fao30 and fao780), health (PubMed) and computer science (Inspec, Krapivin2009, Nguyen2007, SemEval2010, KDD, Wiki20 and WWW). In addition, we constructed another context-specific dataset, KPTimes-Econ, by extracting economy-related news from the KPTimes dataset, which includes 3,258 news articles. For extracting economy-related news articles, we have looked for the records involving the term ``economy" in the \emph{keyword} and/or \emph{categories} field(s). Based on the 11 datasets, we collected a thesaurus (or something similar, e.g., dictionary, ontology, or wordlist) for each context. More specifically, we used the following thesauri: (i) AGROVOC 2021-07~\citep{caracciolo2013agrovoc} -- a multilingual controlled vocabulary constructed by the Food and Agriculture Organization of the United Nations (FAO), with 844,000 agriculture-related terms including 50,163 English ones; (ii) Medical Subject Headings (MeSH) 2021~\citep{lipscomb2000medical} -- a thesaurus covering biomedical and health-related terms produced by the National Library of Medicine (NLM), with over 1.4 million terms in English; (iii) Computer Science Ontology (CSO) v3.3~\citep{salatino2018computer} -- a large-scale computer science ontology automatically produced by Klink-2~\citep{osborne2015klink} algorithm from 16 million computer science publications, with 14,000 terms; and (iv) STW v9.10~\citep{kempf2016role} -- a bilingual thesaurus (in English and German) for economics produced by the Leibniz Information Center for Economics (ZBW), with over 20,000 terms including 6,217 English ones.

For the initial step aiming to experiment with manual integration, we fed each of the baseline methods with each of the datasets and their corresponding thesaurus depending on the context. As in the previous experiment, SIFRank+ was evaluated on only the datasets with shorter documents, i.e., Inspec, KDD, WWW, and KP-Times-Econ in this case. The experiments showed that the manual integration of context-aware thesaurus improved all five AKE methods in terms of precision, recall, and F1-score significantly for all the datasets. The improvement in F1-score was observed to be 29.03\%, 23.88\%, 12.85\%, 13.19\%, and 7.09\% for RaKUn, LexRank, YAKE!, KP-Miner, and SIFRank+, respectively. Table~\ref{tab:lexrank-dict} and Table~\ref{tab:sifrank-dict} show more detailed results of the experiment for LexRank and SIFRank+, respectively. The results of this experiment produced solid evidence of the effectiveness of using context-aware thesaurus to improve the performance of AKE methods.

\begin{table}[htb]
\tabcolsep=0.08cm
\scriptsize
\centering
\caption{Comparison of precision, recall, and F1-score of the original LexRank and its enhanced versions with manual (M) and automatic (A) thesaurus integration, at 10 extracted keywords}
\begin{tabular}{*{11}{c}}
\toprule
\multirow{2}{*}{\textbf{Dataset}} &
\multirow{2}{*}{\textbf{Context}} & \multicolumn{3}{c}{\textbf{LexRank}} & \multicolumn{3}{c}{\textbf{LexRank+T (M)}} &
\multicolumn{3}{c}{\textbf{LexRank+T (A)}}\\
\cmidrule(lr){3-5}\cmidrule(lr){6-8}\cmidrule(lr){9-11}
 & & \textbf{P\%} & \textbf{R\%} & \textbf{F1\%} & \textbf{P\%} & \textbf{R\%} & \textbf{F1\%} & \textbf{P\%} & \textbf{R\%} & \textbf{F1\%}\\
 \midrule
fao30 & Agr. & 20.33 & 6.31 & 9.63 & \textbf{30.33} & \textbf{9.41} & \textbf{14.36} & --- & --- & ---\\
fao780 & Agr. & 8.55 & 10.72 & 9.51 & \textbf{13.04} & \textbf{16.35} & \textbf{14.51} & --- & --- & ---\\
Inspec & CS & 30.49 & 21.61 & 25.29 & \textbf{31.10} & \textbf{22.04} & \textbf{25.79} & 30.97 & 21.95 & 25.69\\
KDD & CS & 6.07 & 14.81 & 8.61 & 6.23 & 15.20 & 8.83 & \textbf{6.25} & \textbf{15.26} & \textbf{8.87}\\
Krapivin2009 & CS & 7.01 & 13.14 & 9.15 & \textbf{8.79} & \textbf{16.48} & \textbf{11.47} & 8.74 & 16.37 & 11.39\\
Nguyen2007 & CS & 13.25 & 11.04 & 12.04 & \textbf{15.69} & \textbf{13.07} & \textbf{14.26} & 15.45 & 12.87 & 14.04\\
SemEval2010 & CS & 13.13 & 8.43 & 10.27 & \textbf{15.10} & \textbf{9.70} & \textbf{11.81} & \textbf{15.10} & \textbf{9.70} & \textbf{11.81}\\
wiki20 & CS & 14.00 & 3.94 & 6.15 & \textbf{23.00} & \textbf{6.48} & \textbf{10.11} & \textbf{23.00} & \textbf{6.48} & \textbf{10.11}\\
WWW & CS & 6.66 & 13.83 & 8.99 & \textbf{6.95} & \textbf{14.43} & \textbf{9.38} & 6.93 & 14.40 & 9.36\\
PubMed & Health & 4.22 & 2.96 & 3.48 & \textbf{8.98} & \textbf{6.31} & \textbf{7.41} & 8.92 & 6.26 & 7.36\\
Schutz2008 & Health & 28.32 & 6.12 & 10.07 & \textbf{34.35} & \textbf{7.43} & \textbf{12.21} & 34.00 & 7.35 & 12.09\\
KPTimes-Econ & Econ. & 3.27 & 7.03 & 4.46 & \textbf{4.09} & \textbf{8.80} & \textbf{5.59} & 4.09 & 8.79 & 5.58\\
\midrule
\textbf{\textit{Avg.~Score (\%)}} &  & 12.94 & 9.99 & 9.80 & \textbf{16.47} & \textbf{12.14} & \textbf{12.14} & 15.35 & 11.94 & 11.63\\
\textbf{\textit{Improvement (\%)}} & & & & & \textcolor{olive}{\textit{27.28}} & \textcolor{olive}{\textit{21.52}} & \textcolor{olive}{\textit{23.88}} & \textcolor{olive}{\textit{21.44}} & \textcolor{olive}{\textit{16.03}} & \textcolor{olive}{\textit{18.07}}\\
\bottomrule
\end{tabular}
\label{tab:lexrank-dict}
\end{table}

\begin{table}[!htb]
\tabcolsep=0.03cm
\scriptsize
\centering
\caption{Comparison of precision, recall, and F1-score of the original SIFRank+ and its enhanced versions with manual (M) and automatic (A) thesaurus integration, at 10 extracted keywords}
\begin{tabular}{*{11}{c}}
\toprule
\multirow{2}{*}{\textbf{Dataset}} &
\multirow{2}{*}{\textbf{Context}} & \multicolumn{3}{c}{\textbf{SIFRank+}} & \multicolumn{3}{c}{\textbf{SIFRank+ + T (M)}} &
\multicolumn{3}{c}{\textbf{SIFRank+ + T (A)}}\\
\cmidrule(lr){3-5}\cmidrule(lr){6-8}\cmidrule(lr){9-11}
& & \textbf{P\%} & \textbf{R\%} & \textbf{F1\%} & \textbf{P\%} & \textbf{R\%} & \textbf{F1\%} & \textbf{P\%} & \textbf{R\%} & \textbf{F1\%}\\
 \midrule
Inspec & CS & 35.68 & 25.29 & 29.60 & \textbf{36.62} & \textbf{25.95} & \textbf{30.37} & 36.03 & 25.53 & 29.88\\
KDD & CS & 5.68 & 13.87 & 8.06 & \textbf{5.97} & \textbf{14.58} & \textbf{8.48} & 5.95 & 14.52 & 8.44\\
WWW & CS & 6.59 & 13.69 & 8.90 & \textbf{7.32} & \textbf{15.19} & \textbf{9.88} & 7.27 & 15.10 & 9.81\\
KPTimes-Econ & Econ. & 3.49 & 7.50 & 4.76 & \textbf{4.56} & \textbf{9.81} & \textbf{6.23} & \textbf{4.56} & \textbf{9.81} & \textbf{6.23}\\
\midrule 
\textbf{\textit{Avg.~Score (\%)}} &  & 12.86 & 15.09 & 12.83 & \textbf{13.62} & \textbf{16.38} & \textbf{13.74} & 13.45 & 16.24 & 13.59\\
\textbf{\textit{Improvement (\%)}} & & & & & \textcolor{olive}{\textit{5.91}} & \textcolor{olive}{\textit{8.55}} & \textcolor{olive}{\textit{7.09}} & \textcolor{olive}{\textit{4.59}} & \textcolor{olive}{\textit{7.62}} & \textcolor{olive}{\textit{5.92}}\\
\bottomrule
\end{tabular}
\label{tab:sifrank-dict}
\end{table}

For the next step, we experimented with the automated thesauri integration process. In our experiments, especially for datasets covering mainly scientific papers, we built a classifier for classifying a given article's title and abstract into the main discipline the article belongs to. The classifier was trained on samples extracted from the arXiv.org dataset\footnote{\url{https://www.kaggle.com/Cornell-University/arxiv}} containing metadata of over 1.7M preprints in multiple disciplines. Before the training process, we filtered the arXiv.org dataset by the main discipline reflected by its \emph{categories} field so as to include the following three disciplines: 1) cs (Computer Science, e.g., cs.AI), 2) bio (Biology, e.g., q-bio), and 3) fin (Finance, e.g., q-fin.CP) and econ (Economics, e.g., econ.EM). After this filtering process, we obtained a dataset of 583,796 samples (551,443 computer science, 20,110 biology, and 12,243 finance/economics samples). Since the resulting dataset is highly imbalanced, we applied random downsampling to equate the number of samples from each discipline to the size of the smallest class, 12,243, which made the final size of our training set 36,729. In our classifier, we utilised the TF-IDF vectoriser for feature extraction. We chose to use the calibrated linear support vector classifier (SVC) with the default parameters and the one-vs-rest setting, rather than a multi-class classification method or more advanced feature extraction methods such as BERT, to show that even a lightweight classifier is sufficient for the task of automatic context detection. The classifier was evaluated with a stratified 5-fold cross-validation. The testing accuracies\footnote{The fraction of the number of correct predictions with respect to the total number of predictions} of computer science, biology and finance/economics models were 93.2\%, 94.9\%, and 97.0\%, respectively. The classifier can also be extended to support multiple contexts for a single article, although in our experiments we considered the case of a single context per article for the sake of simplicity and clarity. We used the Scikit-learn library~\citep{pedregosa2011scikit} to implement all of the mentioned components.

Since the training set of the classifier does not cover agriculture preprints, and we were unable to find a proper agriculture dataset for training, we excluded the agriculture context and the corresponding datasets, fao30 and fao780, for this part of the experiments. The results of the experiments performed with our classifier indicated that the automatic thesaurus integration approach achieved as good as the manual integration approach with a negligible performance decrease. More precisely, the F1-score was improved by an average of 23.23\%, 18.07\%, 9.60\%, 11.27\%, and 5.92\% for RaKUn, LexRank, YAKE!, KP-Miner, and SIFRank+, respectively, compared to the baseline scores. Table~\ref{tab:lexrank-dict} and Table~\ref{tab:sifrank-dict} show more detailed results of the experiment for LexRank and SIFRank+, respectively. The obtained results imply that automatic integration can be generalised to cover more contexts and thesauri, which can be quite useful in real-world AKE applications.

\subsection{Wikipedia Named Entities}

For this part of the experiments, we used the entire set of datasets as we did in Section~\ref{subsec:pos-tag-patterns}. The results of the experiment indicated that leveraging Wikipedia named entities improved the performance of KP-Miner and RaKUn for 16 of the datasets, and the performance of YAKE!\ and LexRank for all the datasets, in terms of all the evaluation metrics. Furthermore, the average improvement rates of the F1-score were observed as 18.83\%, 11.11\%, 10.96\%, and 10.11\% for RaKUn, LexRank, YAKE!, and KP-Miner, respectively. However, we observed a slight decrease in the average F1-score of SIFRank+, although it improved for most (5 out of 7) of the datasets, which may be explained by its underlying sentence embedding approach, SIF~\citep{arora2017a}, which already leverages Wikipedia for pre-training and fine-tuning. Table~\ref{tab:rakun-wiki} and Table~\ref{tab:sifrank-wiki} show more detailed results for RaKUn and SIFRank+ as examples.

\begin{table}[!htb]
\caption{Comparison of precision, recall, and F1-score of the original RaKUn and its enhanced versions with Wikipedia, at 10 extracted keywords}
\scriptsize
\centering
\begin{tabular}{*{7}{c}}
\toprule
\multirow{2}{*}{\textbf{Dataset}} &  \multicolumn{3}{c}{\textbf{RaKUn}} & \multicolumn{3}{c}{\textbf{RaKUn+Wiki}}\\
\cmidrule(lr){2-4}\cmidrule(lr){5-7}
& \textbf{P\%} & \textbf{R\%} & \textbf{F1\%} &  \textbf{P\%} & \textbf{R\%} & \textbf{F1\%}\\
\midrule
KPCrowd & 42.52 & 8.64 & 14.36 & \textbf{42.64} & \textbf{8.66} & \textbf{14.40}\\
citeulike180 & 16.56 & 9.50 & 12.08 & \textbf{17.92} & \textbf{10.29} & \textbf{13.07}\\
DUC-2001 & 5.68 & 7.03 & 6.29 & \textbf{6.17} & \textbf{7.64} & \textbf{6.82}\\
fao30 & 15.00 & 4.65 & 7.10 & \textbf{18.67} & \textbf{5.79} & \textbf{8.84}\\
fao780 & 6.50 & 8.14 & 7.23 & \textbf{7.64} & \textbf{9.57} & \textbf{8.50}\\
Inspec & 6.54 & 4.64 & 5.43 & \textbf{6.74} & \textbf{4.77} & \textbf{5.59}\\
KDD & \textbf{3.66} & \textbf{8.92} & \textbf{5.19} & 3.63 & 8.86 & 5.15\\
KPTimes & 8.07 & 16.03 & 10.74 & \textbf{8.15} & \textbf{16.18} & \textbf{10.84}\\
Krapivin2009 & 2.77 & 5.20 & 3.62 & \textbf{4.94} & \textbf{9.26} & \textbf{6.44}\\
Nguyen2007 & 6.79 & 5.66 & 6.17 & \textbf{9.67} & \textbf{8.05} & \textbf{8.78}\\
PubMed & 4.30 & 3.02 & 3.55 & \textbf{6.58} & \textbf{4.62} & \textbf{5.43}\\
Schutz2008 & 33.14 & 7.16 & 11.78 & \textbf{40.09} & \textbf{8.67} & \textbf{14.25}\\
SemEval2010 & 6.75 & 4.33 & 5.28 & \textbf{10.04} & \textbf{6.45} & \textbf{7.85}\\
SemEval2017 & 11.42 & 6.60 & 8.37 & \textbf{11.74} & \textbf{6.79} & \textbf{8.60}\\
theses100 & 3.90 & 5.85 & 4.68 & \textbf{4.80} & \textbf{7.20} & \textbf{5.76}\\
wiki20 & 9.50 & 2.68 & 4.18 & \textbf{19.50} & \textbf{5.49} & \textbf{8.57}\\
WWW & 4.32 & 8.98 & 5.84 & \textbf{4.39} & \textbf{9.12} & \textbf{5.93}\\
\midrule
\textit{\textbf{Avg.~Score (\%)}} & 11.02 & 6.88 & 7.17 & \textbf{13.14} & \textbf{8.08} & \textbf{8.52}\\
\textit{\textbf{Improvement (\%)}} &  &  & & \textcolor{olive}{\textit{19.24}} & \textcolor{olive}{\textit{17.44}} & \textcolor{olive}{\textit{18.83}}\\
\bottomrule
\end{tabular}
\label{tab:rakun-wiki}
\end{table}

\begin{table}[!htb]
\caption{Comparison of the precision, recall, and F1-score of the original SIFRank+ and the one utilising Wikipedia named entities, at 10 extracted keywords}
\scriptsize
\centering
\begin{tabular}{*{7}{c}}
\toprule
\multirow{2}{*}{\textbf{Dataset}} & \multicolumn{3}{c}{\textbf{SIFRank+}} & \multicolumn{3}{c}{\textbf{SIFRank+ + Wiki}}\\
\cmidrule(lr){2-4}\cmidrule(lr){5-7}
& \textbf{P\%} & \textbf{R\%} & \textbf{F1\%} & \textbf{P\%} & \textbf{R\%} & \textbf{F1\%}\\
\midrule
KPCrowd & 26.08 & 5.30 & 8.81 & \textbf{27.46} & \textbf{5.58} & \textbf{9.27}\\
DUC-2001 & \textbf{28.34} & \textbf{35.09} & \textbf{31.36} & 22.82 & 28.26 & 25.25\\
Inspec & 35.68 & 25.29 & 29.60 & \textbf{36.60} & \textbf{25.94} & \textbf{30.36}\\
KDD & 5.68 & 13.87 & 8.06 & \textbf{6.11} & \textbf{14.90} & \textbf{8.66}\\
KPTimes & 7.92 & 15.74 & 10.54 & \textbf{9.22} & \textbf{18.31} & \textbf{12.26}\\
SemEval2017 & \textbf{41.66} & \textbf{24.08} & \textbf{30.52} & 41.34 & 23.89 & 30.28\\
WWW & 6.59 & 13.69 & 8.90 & \textbf{7.50} & \textbf{15.57} & \textbf{10.12}\\
\midrule
\textbf{\textit{Avg.~Score (\%)}} & \textbf{21.71} & \textbf{19.01} & \textbf{18.26} & 21.58 & 18.92 & 18.03\\
\textbf{\textit{Improvement (\%)}} & & & & \textcolor{red}{\textit{-0.60}} & \textcolor{red}{\textit{-0.47}} & \textcolor{red}{\textit{-1.26}}\\
\bottomrule
\end{tabular}
\label{tab:sifrank-wiki}
\end{table}

\subsection{Combining Post-Processing Steps}

\begin{figure*}[htb!]
\centering
\subcaptionbox{YAKE!}{\includegraphics[width=0.48\linewidth]{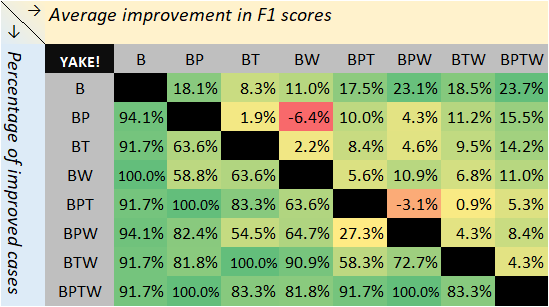}}
~
\subcaptionbox{KP-Miner}{\includegraphics[width=0.48\linewidth]{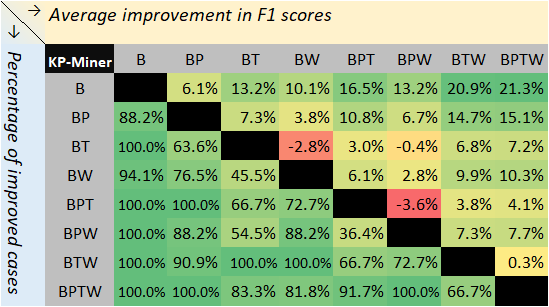}}
~
\subcaptionbox{RaKUn}{\includegraphics[width=0.48\linewidth]{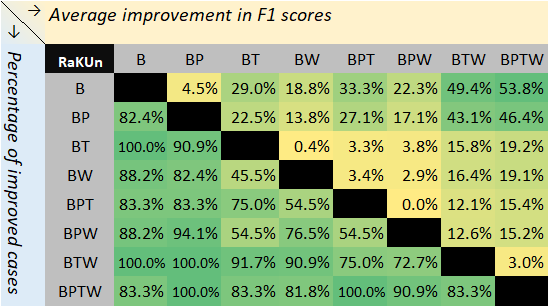}}
~
\subcaptionbox{LexRank}{\includegraphics[width=0.48\linewidth]{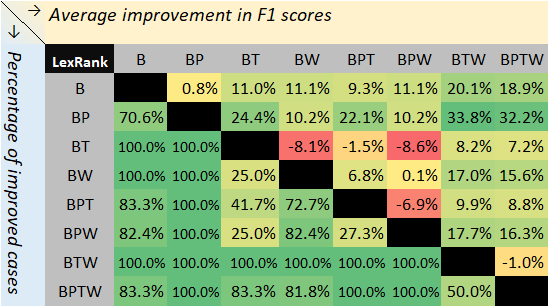}}
~
\subcaptionbox{SIFRank+}{\includegraphics[width=0.48\linewidth]{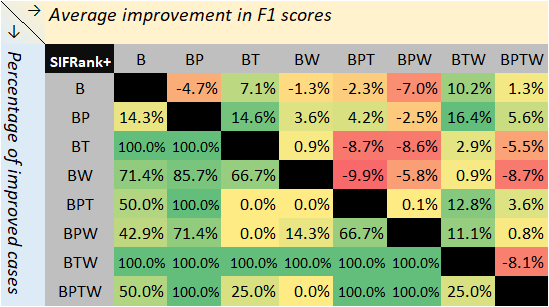}}
\caption{Average improvements in F1 scores across all the datasets (upper side), and percentages of the improved cases across all the datasets (bottom side), for different AKE methods. (B: Baseline, P: PoS-tagging, T: Thesaurus integration, W: Wikipedia integration)}
\label{fig:heatmap}
\end{figure*}

In the final part of our experiments, we tried combining multiple post-processing steps to improve the performance further. With this respect, we tried to apply all the combinations of the three proposed enhancements. The generated heatmaps from the F1@10 scores and the percentages of improved cases with different combinations for each baseline method can be seen in Figure~\ref{fig:heatmap}. The results show that the best F1 scores for YAKE!, RaKUn, and KP-Miner were obtained when all the proposed post-processing steps were applied. For LexRank and SIFRank+, however, the best combination was integrating context-aware thesaurus and Wikipedia since they already benefited from PoS-tagging-based filtering. In addition, the applied post-processing steps improved the baselines significantly -- the improvement rate reached up to 23.7\% for YAKE!, 21.3\% for KP-Miner, 53.8\% for RaKUn, 20.1\% for LexRank, and 10.2\% for SIFRank+. Finally, the improvements were consistent -- at least one combination of the post-processing steps was observed for each method, resulting in higher performance across all the datasets. The results showed that even for more modern AKE methods there is still room for improvement using simple post-processing steps like those proposed in this paper.

\section{Further Discussions}
\label{sec:discussion}

The proposed post-processing steps in this study were applied to five representative SOTA AKE methods, showing their universality to improve the performance of many different AKE methods. The universality of the post-processing steps is rooted in the fact that they rely on access to the list of candidate keywords and their scores, which are the standard output for most (if not all) AKE methods. The performance improvements can be explained by two main reasons: (i) utilising PoS-tagging avoids AKE methods, especially those less benefiting from linguistic features, to generate keywords that are less likely to be meaningful keywords, such as conjunctions, determiners, and adverbs; and (ii) thesauri and Wikipedia-based enhancements allow prioritisation of more domain-specific and context-specific keywords to be returned by AKE methods.

Although PoS-tagging can be easily integrated into AKE methods to implement a filtering mechanism, it should be separately considered for each dataset since AKE datasets lack linguistic standards for golden keywords. This can significantly increase the accuracy of the AKE methods benefiting from PoS-tagging. Thesaurus and Wikipedia integration can also be applied to AKE methods without much effort. Considering that a text document can cover multiple contexts, the results we reported can be further improved by integrating multiple contexts. This can be achieved by utilising a multi-label classifier. Since one-vs-rest classifiers can be used for multi-label classification, our classifier can be refined to cover multiple contexts. In addition, more advanced models, such as BERT, can be utilised to develop a more accurate classifier. It is also worth noting that two of the proposed post-processing steps in this study were selected as representative examples of semantic elements. Other semantic elements can also be used to further improve the performance of AKE methods.

Although our experiments on the proposed post-processing steps are based on English NLP tools and datasets, they can also be applied to multilingual AKE methods, e.g., YAKE!, for any language. The language of input documents can be identified automatically with a language identifier, which can achieve high accuracy for many languages~\citep{jauhiainen2019automatic}. Then, the corresponding PoS-tagger and Wikipedia data can be utilised, although the set of acceptable PoS-tag patterns will need updating according to the identified language. Nevertheless, utilising a context-aware thesaurus could be tricky for some languages especially small ones as there might be no thesaurus relevant to the context of the document in the identified language.

This study has a number of limitations that can be addressed in future work. Firstly, the selected baseline AKE methods are just examples of SOTA methods so may not be sufficiently representative. As our focus was improving AKE methods in general, we did not aim to achieve the best scores among the studies on AKE. As a result, this study is limited to open-source, unsupervised, and general-purpose AKE methods. In addition, this study leveraged multiple elements of the English language and used English datasets for evaluation. Therefore, it disregarded non-English settings, which are needed especially for multilingual AKE methods, such as YAKE!. Besides, the proposed mechanisms have been applied separately throughout the experiments. Therefore, the results could be improved further if different mechanisms benefit from each other (e.g., applying PoS-tag-based filtering to the Wikipedia integration mechanism to disregard the Wikipedia named entities that cannot be a keyword). Finally, a better matching strategy considering word ambiguities can be developed for checking if a candidate keyword appears in a thesaurus or Wikipedia, with the help of techniques such as word sense disambiguation.

\section{Conclusion}
\label{sec:conclusion}

AKE has a more important role in IR and NLP with the increasingly vast amount of digital textual data that modern systems process. In this paper, we aimed to show that an enhanced level of semantic-awareness supported by PoS-tagging can improve AKE algorithms. We selected five algorithms as the baseline methods upon experiments comparing several state-of-the-art AKE methods. Then, we used PoS-tagging, integrated thesauri, and Wikipedia named entities for improving the baselines. Our experiments on 17 English datasets indicated that the three proposed mechanisms improved the baseline algorithms significantly and consistently.

\subsubsection*{Acknowledgements}

We would like to thank Ricardo Campos for clarification and additional information about the YAKE!\ algorithm. The first author E.~Altuncu was supported by funding from the Ministry of National Education, Republic of Türkiye, under grant number MoNE-YLSY-2018.

\section*{Statements and Declarations}

\subsubsection*{Competing Interests}

The authors declare no competing interest. No funding was received for conducting this
study.

\subsubsection*{Data Availability}

The source code and the complete experimental results of our work will be released after the paper is accepted for publication.

\bibliography{main}

\end{document}